\documentclass[a4paper, 10pt, conference]{ieeeconf}      % Use this line for a4 paper

\usepackage{cite}
\usepackage{amsmath,amssymb,amsfonts}
\usepackage{graphicx}
\usepackage{textcomp}
\usepackage{booktabs}          
\usepackage{makecell}   
\usepackage{tabularx}
\usepackage{float} 
\usepackage{textcomp}
\usepackage{xcolor}
\usepackage{xcolor}
\usepackage{algorithm}
\usepackage{algpseudocode}

\counterwithout{table}{section}
\usepackage[hidelinks]{hyperref}
\title{\LARGE \bf
Loop Closure using AnyLoc Visual Place Recognition in DPV-SLAM
}

\author{
\authorblockN{
Wenzheng Zhang\authorrefmark{1}, Kazuki Adachi\authorrefmark{1}, Yoshitaka Hara\authorrefmark{2},
Sousuke Nakamura\authorrefmark{3}
}
\authorblockA{
\\
\begin{minipage}[c]{0.35\hsize}
\centering
\authorrefmark{1}
Graduate School of Science and\\ Engineering, Hosei University, \\Tokyo, Japan
\end{minipage}%
\begin{minipage}[c]{0.35\hsize}
\centering
\authorrefmark{2}
    Future Robotics Technology Center (fuRo), Chiba Institute of Technology, \\Chiba, Japan
\end{minipage}%
\begin{minipage}[c]{0.35\hsize}
\centering
\authorrefmark{3}
 Faculty of Science and\\ Engineering, Hosei University, \\Tokyo, Japan
\end{minipage}%
}
}

\begin{document}
\maketitle
\thispagestyle{empty}
\pagestyle{empty}

%%%%%%%%%%%%%%%%%%%%%%%%%%%%%%%%%%%%%%%%%%%%%%%%%%%%%%%%%%%%%%%%%%%%%%%%%%%%%%%%
\begin{abstract}
Loop closure is crucial for maintaining the accuracy and consistency of visual SLAM. We propose a method to improve loop closure performance in DPV-SLAM. Our approach integrates AnyLoc, a learning-based visual place recognition technique, as a replacement for the classical Bag of Visual Words (BoVW) loop detection method. In contrast to BoVW, which relies on handcrafted features, AnyLoc utilizes deep feature representations, enabling more robust image retrieval across diverse viewpoints and lighting conditions. Furthermore, we propose an adaptive mechanism that dynamically adjusts similarity threshold based on environmental conditions, removing the need for manual tuning. Experiments on both indoor and outdoor datasets demonstrate that our method significantly outperforms the original DPV-SLAM in terms of loop closure accuracy and robustness. The proposed method offers a practical and scalable solution for enhancing loop closure performance in modern SLAM systems.
\end{abstract}

%%%%%%%%%%%%%%%%%%%%%%%%%%%%%%%%%%%%%%%%%%%%%%%%%%%%%%%%%%%%%%%%%%%%%%%%%%%%%%%%
\section{Introduction}
% 同时定位与建图（SLAM） 是实现移动机器人系统自主导航的基础性技术。在众多 SLAM 模态中，视觉 SLAM（Visual SLAM） 仅依赖视觉输入（通常来自单目或双目相机）实现自我定位与地图构建。
Simultaneous Localization and Mapping (SLAM) is a fundamental technology that enables autonomous navigation in mobile robotic systems. Among the various SLAM modalities, visual SLAM achieves both robot localization and map building solely based on visual input, typically from monocular or stereo cameras.

% 在单目视觉 SLAM中，由于缺乏深度信息，随着时间推移通常会出现累计误差和尺度不一致的问题，这会影响地图的整体一致性。回环检测（loop closure） 是一种关键机制，通过识别机器人先前访问过的位置并对地图进行调整，从而修正这些不一致。
% 在现实环境中，移动机器人常常面临光照变化、动态物体遮挡、视角变化等复杂情况，这些都对视觉回环检测提出了极高要求。传统的 BoVW 方法在面对这些变化时，往往容易出现误匹配或漏检，导致地图构建的精度下降。
In monocular visual SLAM, accumulated drift and scale inconsistency inevitably emerge over time, potentially compromising the global consistency of the constructed map. Loop closure is a crucial process for mitigating these issues by recognizing previously visited locations and subsequently refining the map.

In real-world environments, mobile robots often face challenging conditions such as illumination changes, dynamic occlusions, and substantial viewpoint variations. These factors impose strict demands on the robustness of visual loop detection. Conventional Bag of Visual Words (BoVW) approaches are particularly vulnerable under such circumstances, frequently resulting in false matches or missed detections, and thereby degrading both the accuracy and consistency of SLAM systems.

% 近年来，基于深度学习的视觉特征表示取得了显著进展，为回环检测提供了新的方法，超越了传统的“视觉词袋模型”（BoVW）。例如，AnyLoc \cite{Keetha_2023} 是一种视觉位置识别方法，通过预训练获得了较强的泛化能力和语义区分能力，
% 与传统 BoVW 方法依赖局部特征（如 ORB、SIFT）和手工构建的词袋模型不同，AnyLoc 基于深度卷积神经网络，能够从图像中提取更加稳健且具有语义区分性的特征表示。其预训练机制使得模型在跨场景、跨时间的条件下仍保持较强的回环识别能力。
Recent advances in learning-based visual feature representations have enabled novel loop detection methods that surpass classical approaches such as BoVW. In particular, AnyLoc \cite{Keetha_2023}, a visual place recognition technique, achieves high generalization and semantic discrimination through extensive pretraining, demonstrating superior robustness to environmental variations compared to conventional methods. 

Unlike classical BoVW approaches that depend on handcrafted local features such as ORB\cite{orb} or SIFT \cite{lowe2004distinctive} and manually constructed vocabularies, AnyLoc utilizes deep neural networks to extract robust and semantically meaningful image representations. Its pretraining strategy facilitates strong generalization across diverse environments and temporal changes, making it highly suitable for loop detection in real-world environments.

% 在我们以往的研究工作中 \cite{Adachi2025}，我们评估了多种单目视觉 SLAM 系统，发现 DPV-SLAM \cite{Lipson2025DeepPatchVisualSLAM} 在各方面表现出色。PS:DPV-SLAM 是一种基于 Deep Patch Visual Odometry (DPVO) 扩展的单目视觉 SLAM 方法，通过高效利用深度特征实现实时运行和较低显存占用。DPV-SLAM 虽然 DPV-SLAM 在视觉里程计模块中引入了深度学习方法，但其回环检测仍然采用传统的 BoVW 方法。
In our previous work \cite{Adachi2025}, we evaluated several monocular visual SLAM systems and found that Deep Patch Visual SLAM (DPV-SLAM) \cite{Lipson2025DeepPatchVisualSLAM} demonstrated outstanding performance. DPV-SLAM is a monocular visual SLAM method extended from Deep Patch Visual Odometry (DPVO) \cite{teed2023deep}, which efficiently leverages deep feature representations to achieve real-time performance with low memory consumption. While DPV-SLAM leverages deep learning in its visual odometry module, it still relies on the classical BoVW-based approach for loop detection.

In this paper, we propose to enhance the loop closure module of DPV-SLAM by replacing its BoVW-based loop detection with AnyLoc. We further introduce a complete loop closure pipeline that combines adaptive similarity thresholding with geometric verification. The proposed method is evaluated using real-world camera data to validate its effectiveness.

Our contributions are as follows: 
% 本文旨在通过将回环检测模块从 BoVW 替换为 AnyLoc 来增强 DPV-SLAM 的回环检测能力。我们还提出了一个完整的回环检测流程，结合了自适应相似度阈值调整机制与几何验证策略。我们使用真实世界的摄像头数据对该方法进行了评估，以验证其有效性。我们的主要贡献如下：
\begin{itemize}
 \item We integrate AnyLoc, a learning-based visual place recognition method, into loop closure of DPV-SLAM.
 \item We introduce an automatic threshold adjustment mechanism that adapts the similarity threshold to different environments.
 \item Experiments show that our method outperforms the original DPV-SLAM in terms of loop closure accuracy and robustness.
\end{itemize}
% 将基于深度学习的视觉位置识别方法 AnyLoc 集成至 DPV-SLAM 中，以提升回环检测的准确性与鲁棒性；
% 提出一种自动阈值调整机制，能够根据不同环境自适应地设定相似度阈值。
% PS
\begin{figure*}[tb]
    \centering
    \includegraphics[width=1\linewidth]{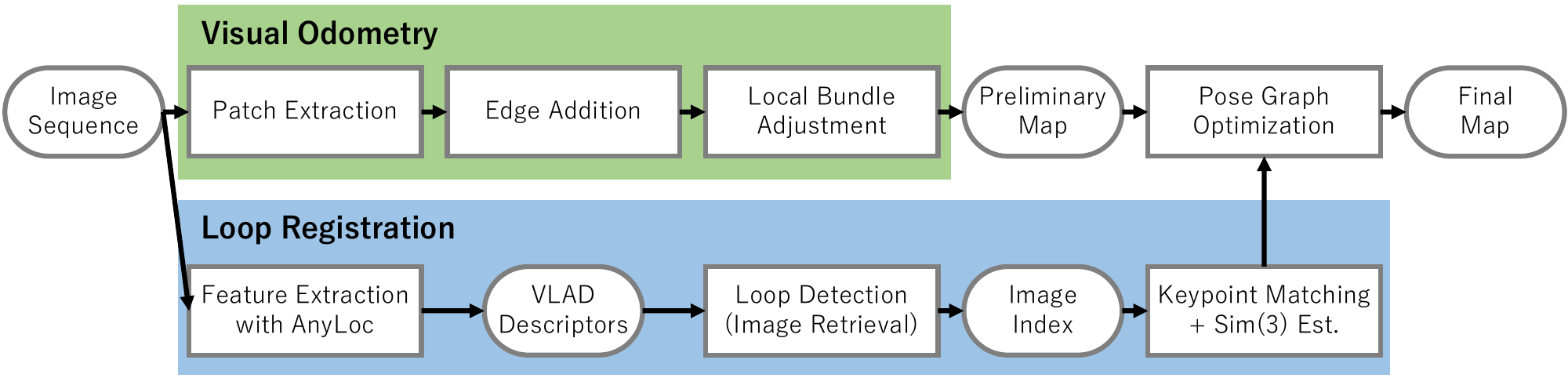}
    \caption{DPV-SLAM pipeline with AnyLoc-based visual place recognition.}
    \label{fig:Pipeline}
\end{figure*}
\section{Related Work}
Research related to the proposed method can be broadly categorized into three groups: loop detection methods based on Bag of Visual Words (BoVW), loop detection methods utilizing learning-based visual features, and loop detection methods leveraging topological and semantic information.

In many conventional visual SLAM systems, loop detection commonly relies on BoVW approaches such as DBoW2 \cite{GalvezTRO12}. These methods quantize local features like ORB or SIFT into visual words and enable fast image retrieval based on word frequency, achieving high computational efficiency and processing speed. Such approaches have been adopted in systems like ORB-SLAM \cite{Mur_Artal_2015} and DPV-SLAM. However, BoVW-based methods often suffer from performance degradation caused by illumination changes and viewpoint variations.

To address these challenges, loop detection methods using deep learning have attracted attention. Techniques like NetVLAD \cite{Arandjelovic2016NetVLAD} and Patch-NetVLAD \cite{Hausler2021PatchNetVLAD} aggregate local descriptors into global image descriptors, enabling robust visual place recognition. Among these, AnyLoc is trained on large-scale datasets covering scenes across multiple cities, achieving high generalization capability and semantically meaningful feature representations in both structured and unstructured environments.

Furthermore, topological loop detection methods that exploit spatial structure and temporal context have also been proposed. For instance, Topomap \cite{Blochliger2017Topomap} captures the spatial structure of environments using graph representations, contributing to loop detection and map consistency. Similarly, the integration of semantic information as demonstrated in SemanticFusion \cite{McCormac2017SemanticFusion} has proven effective in stabilizing recognition under complex environmental conditions.

In summary, existing SLAM systems have gradually evolved from classical BoVW approaches toward learning-based methods for loop detection. 

Motivated by this trend, this paper proposes the integration of the deep visual place recognition method AnyLoc into the original DPV-SLAM framework, replacing its BoVW module to enhance the robustness and accuracy of loop detection. Furthermore, an adaptive thresholding mechanism is introduced, enabling the system to dynamically adjust image matching threshold in response to environmental changes. This design allows the system to maintain stable performance across diverse and challenging scenarios.
% 综上所述，现有 SLAM 系统在闭环检测中逐渐从传统 BoVW 方法向深度学习驱动的方法演进。基于 此背景，本文提出将深度视觉位置识别方法 Any-Loc 集成至原始的 DPV-SLAM 框架中，以替代其 原有的 BoW 模块，从而提升闭环检测的鲁棒性与准确率。此外，本文进一步设计了一种自适应阈值机制，使得系统能够根据环境变化动态调整图像匹 配的判断标准，从而在多样化场景中保持更稳定的性能表现

\section{Loop Closure with AnyLoc}

\subsection{Overview}
Fig.~\ref{fig:Pipeline} illustrates the pipeline of the proposed method. In the Visual Odometry module, the input image undergoes feature extraction and pose estimation, followed by bundle adjustment to construct a map prior to loop closure.

Running concurrently with this process is the Loop Registration module, which determines the occurrence of loop closures. This module extracts global descriptors using AnyLoc and retrieves multiple candidate images with high similarity scores. Based on a threshold, it decides whether a loop closure has occurred and performs geometric verification. Upon successful verification, the map is optimized through pose graph optimization, generating the final consistent map. 

\subsection{Global Descriptor Extraction}
\begin{figure}[t]
\centering
\includegraphics[width=1\linewidth]{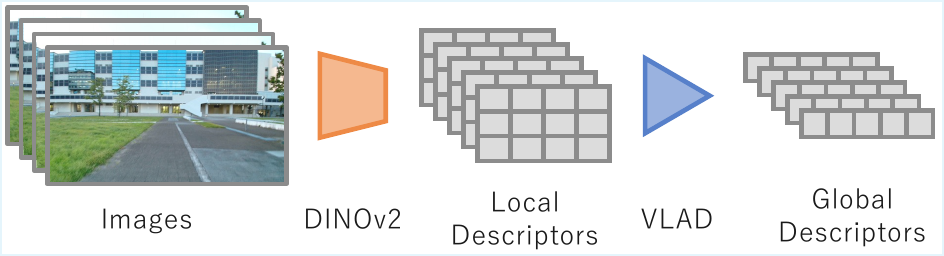}
\caption{Extraction of global descriptors using AnyLoc-VLAD-DINOv2.}
\label{fig:discriptor}
\end{figure}

Fig.~\ref{fig:discriptor} illustrates the extraction of global descriptors. Global descriptors are extracted from keyframe images using AnyLoc-VLAD-DINOv2. In this process, DINOv2 functions as the backbone for visual feature extraction, producing local descriptors from the input images. These local descriptors are then aggregated into highly discriminative global descriptors using the VLAD feature aggregation method. The extracted global descriptors are stored in a database for subsequent retrieval.

\subsection{Adaptive Similarity Threshold Adjustment}
In conventional methods such as BoVW-based AnyLoc, the similarity threshold for loop closure decision must be manually set, which affects the system’s generalization performance. To address this, we propose an adaptive method for similarity threshold adjustment.
\begin{figure}[t]
    \centering
    \includegraphics[width=0.95\linewidth]{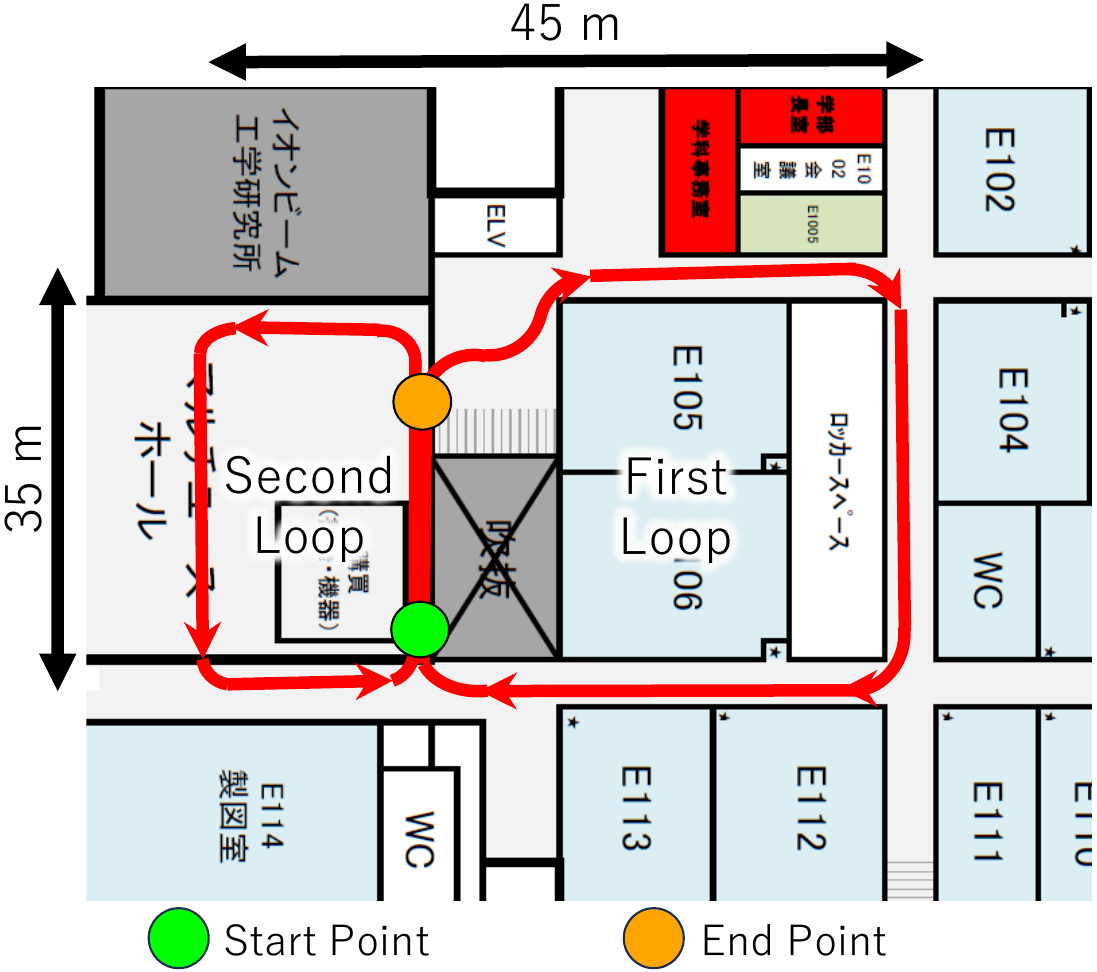}
    \caption{Corridor environment.}
    \label{fig:environment_indoor}
\end{figure}
% 但是由于环境刚开始的不稳定已经之后环境变动的影响，在之后检测到新的threshold之后，threshold将会记录此时的最高的threshold，并将之前的threshold进行更新。从而提高the adaptability and robustness of loop detection.
% Specifically, during the initial phase, a small set of valid loop closure candidate similarity scores (default: 5) are collected, and their median value is calculated and used as the current threshold (loop\_thresh). This threshold reflects the score distribution according to the environment, thereby improving the adaptability and robustness of loop detection.

Specifically, during the initial phase, a small set of valid loop closure candidate similarity scores (default: 5) are collected, and their median value is calculated and used as the current threshold (loop\_thresh). This threshold reflects the score distribution according to the environment. 

Nevertheless, owing to the initial instability of the environment and the influence of subsequent environmental variations, once a new threshold is detected, the system records the maximum threshold observed at that moment and updates the previous value accordingly. This updating mechanism improves the adaptability and robustness of the loop detection process.

\subsection{Loop Detection and Sim(3) Pose Registration}
After the threshold is determined, the similarity between the global descriptor of each keyframe and the query is computed, and keyframes exceeding the loop\_thresh are considered as loop candidates. Temporally close frames (e.g., within 50 frames) are excluded, and the candidate with the highest similarity is selected.

Following loop detection, DISK\cite{tyszkiewicz2020disk} keypoints are extracted and matched using LightGlue\cite{lindenberger2023lightglue}. Then, a Sim(3) transformation is estimated via RANSAC\cite{ransac} combined with the Umeyama algorithm\cite{umeyama}. Geometric verification is considered successful if the number of inliers exceeds a predefined  threshold (30 points), and the estimated pose is incorporated into the pose graph as a loop constraint.

\subsection{Pose Graph Optimization}
% DPV_SLAM 在成功完成回环检测与几何验证后，估计得到的 Sim(3) 变换被加入到位姿图中，作为回环约束。我们利用 DPV-SLAM 中的 CUDA 加速块稀疏优化后端（block-sparse optimization backend） 来执行非线性的位姿图优化（PGO），从而实现高效的实时全局位姿精炼。该优化过程在独立线程中异步运行，优化完成后通过回调函数更新整条轨迹，从而实现在线地图的持续优化。
After successful loop detection and geometric verification, the estimated Sim(3) transformation is added into the pose graph as a loop closure constraint. We utilize the CUDA-accelerated block-sparse optimization backend from DPV-SLAM to perform nonlinear pose graph optimization (PGO), achieving efficient and real-time global pose refinement. The optimization runs asynchronously in a separate thread, and upon completion, a callback function updates the entire trajectory, enabling continuous online map optimization.

% PS 
\begin{figure}[t]
    \centering
    \includegraphics[width=1 \linewidth]{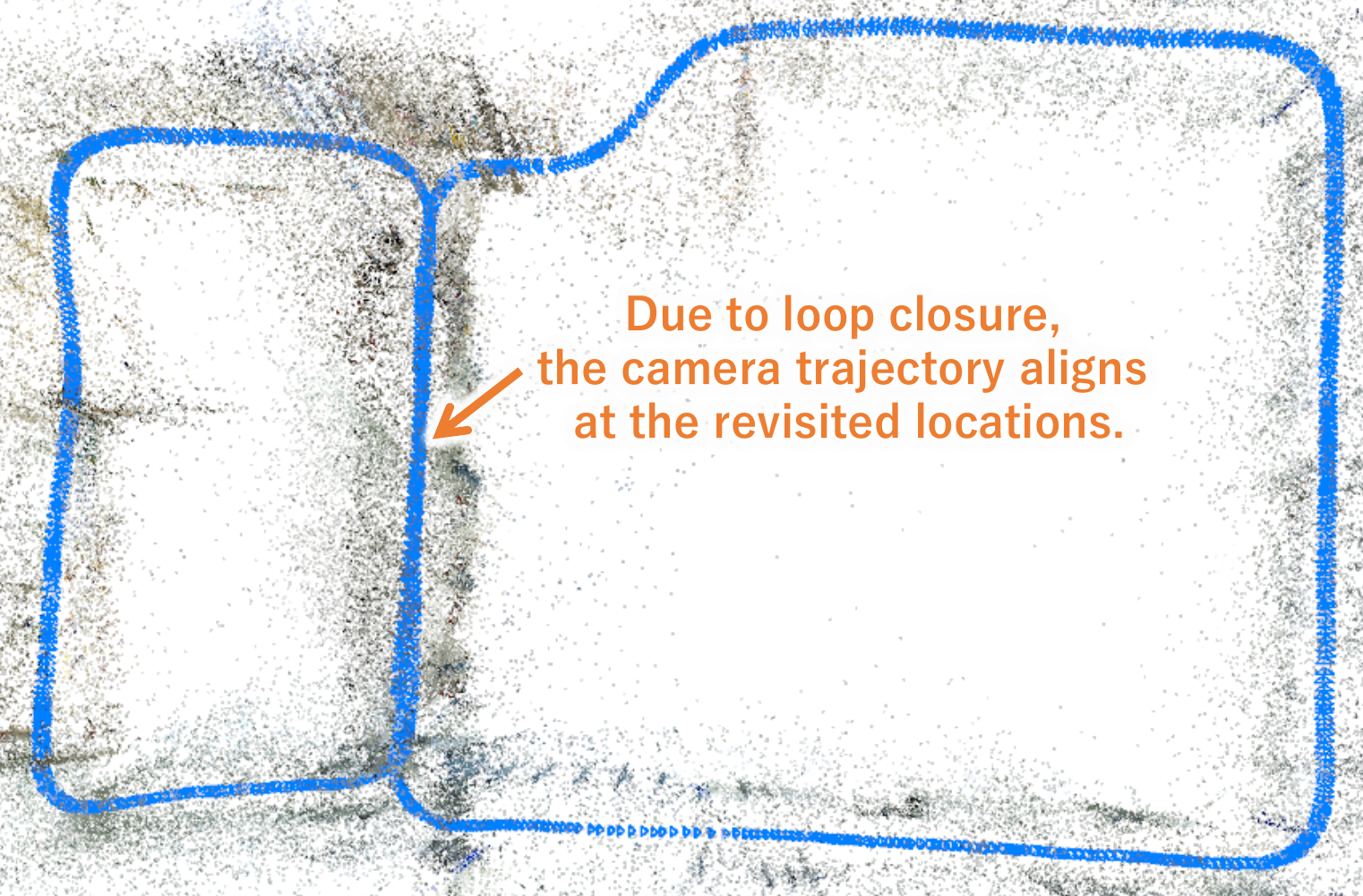}
    \caption{3D point cloud map and camera trajectory in the corridor environment (proposed method).}
    \label{fig:experiment_indoor}
\end{figure}
% PS
\begin{figure}[t]
    \centering
    \includegraphics[width=1\linewidth]{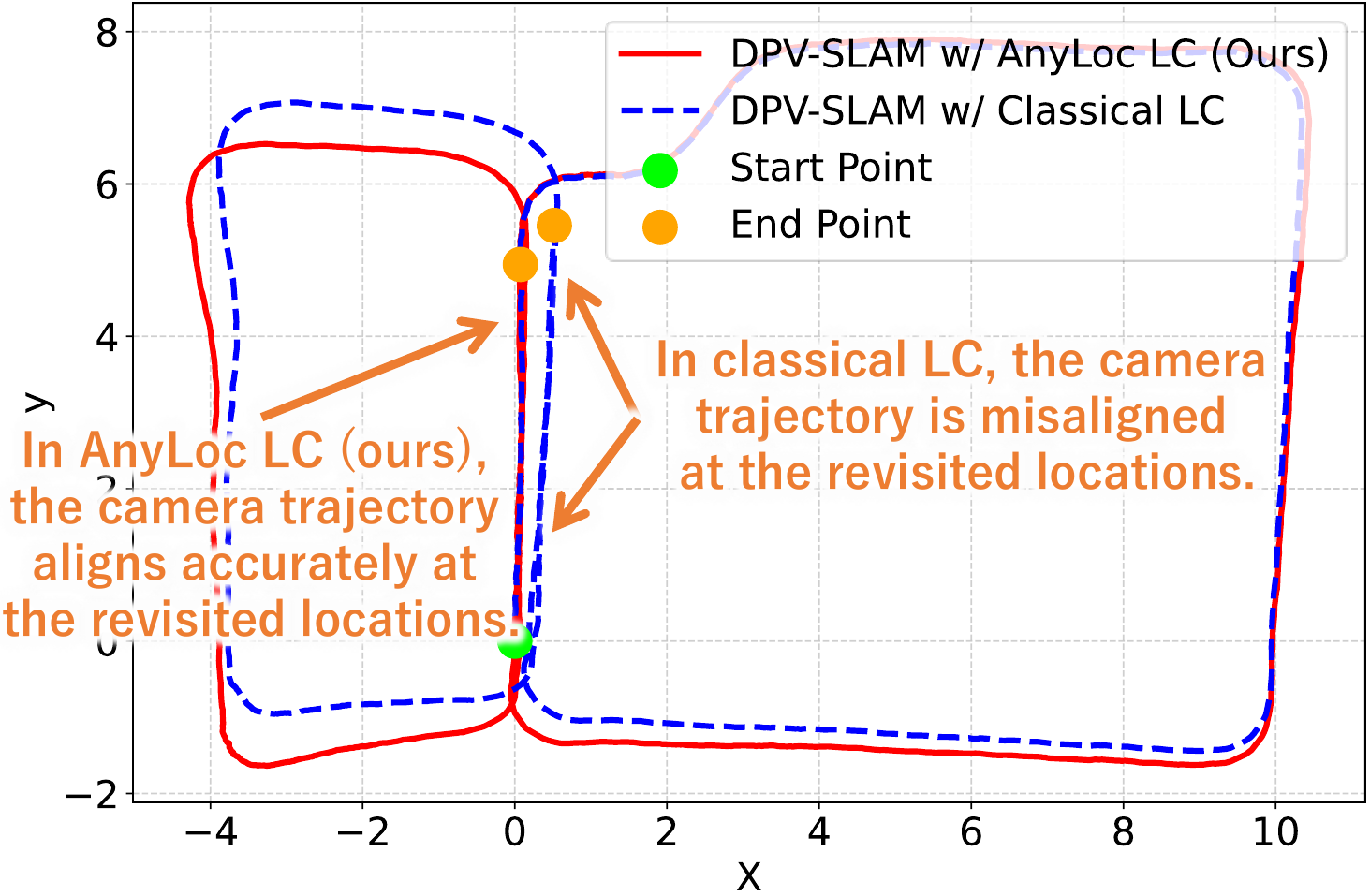}
    \caption{Comparison of camera trajectories in the corridor environment.}
    \label{fig:comparison_indoor}
\end{figure}

\section{Experiments}
\subsection{Overview}
We conducted experiments in several real-world environments and compared our approach with previous work. Specifically, this study  aims to answer the following questions:\\
\noindent\textbf{Q1:} \label{q1} Does the proposed method improve the accuracy of loop detection compared to the baseline, thereby contributing to a more accurate map? \\
%我们的方法和原来的方法相比，是否有效地提高了闭环检测的准确率，从而使得得到的地图更加准确？
\noindent\textbf{Q2:} \label{q2}
Does the proposed method exhibit better generalization capability than previous approach across diverse environments?
% 我们的方法是否比原来的方法的泛用性更高？
\subsection{Environments}
Experiments were conducted in both indoor and outdoor environments at the Koganei Campus of Hosei University. 

Fig.~\ref{fig:environment_indoor} shows the indoor environment and its trajectory. The indoor environment in Fig.~\ref{fig:environment_indoor} is relatively dark and features multiple occurrences of pedestrian crossings. 
% PS 3 Images
\begin{figure}[t]
    \centering
    \includegraphics[width=0.85\linewidth]{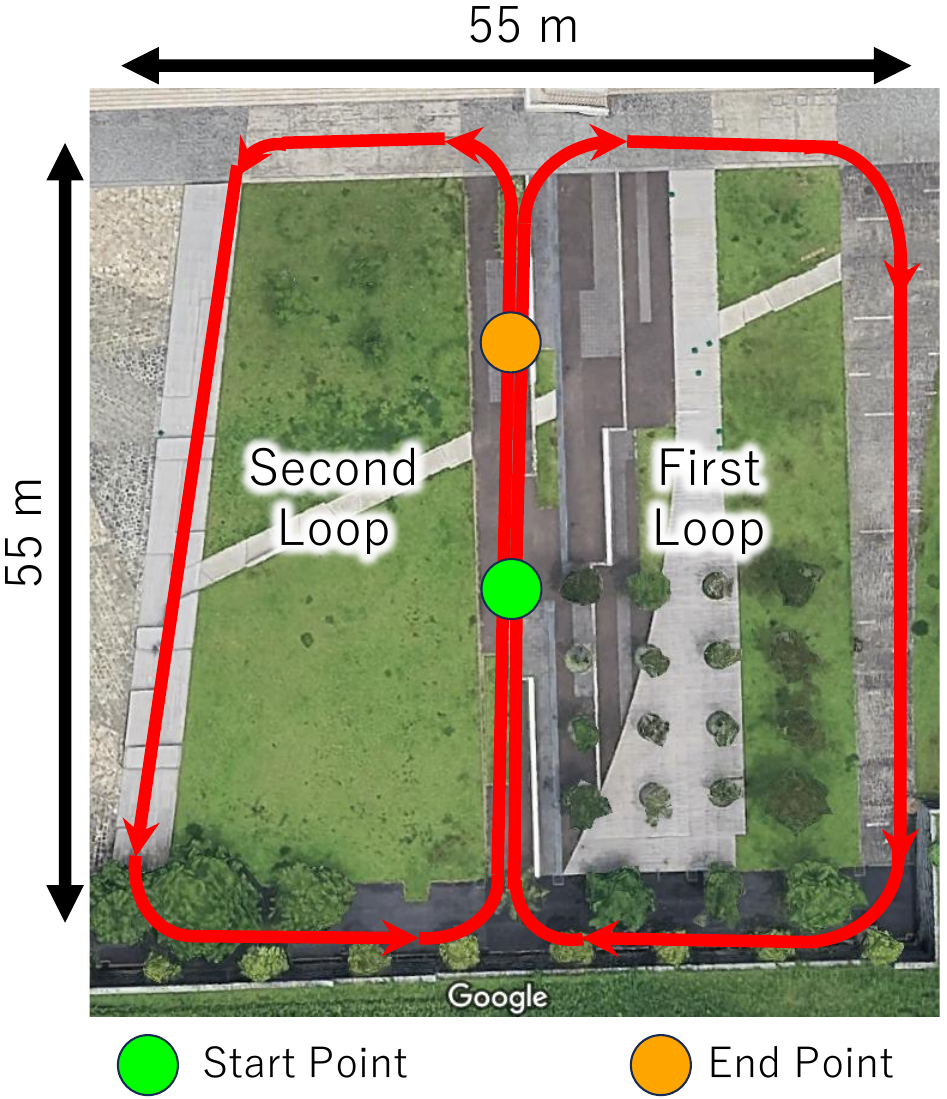}
    \caption{Outdoor environment (courtyard).}
    \label{fig:environment_outdoor2}
\end{figure}
\begin{figure}[t]
    \centering
    \includegraphics[width=0.8\linewidth]{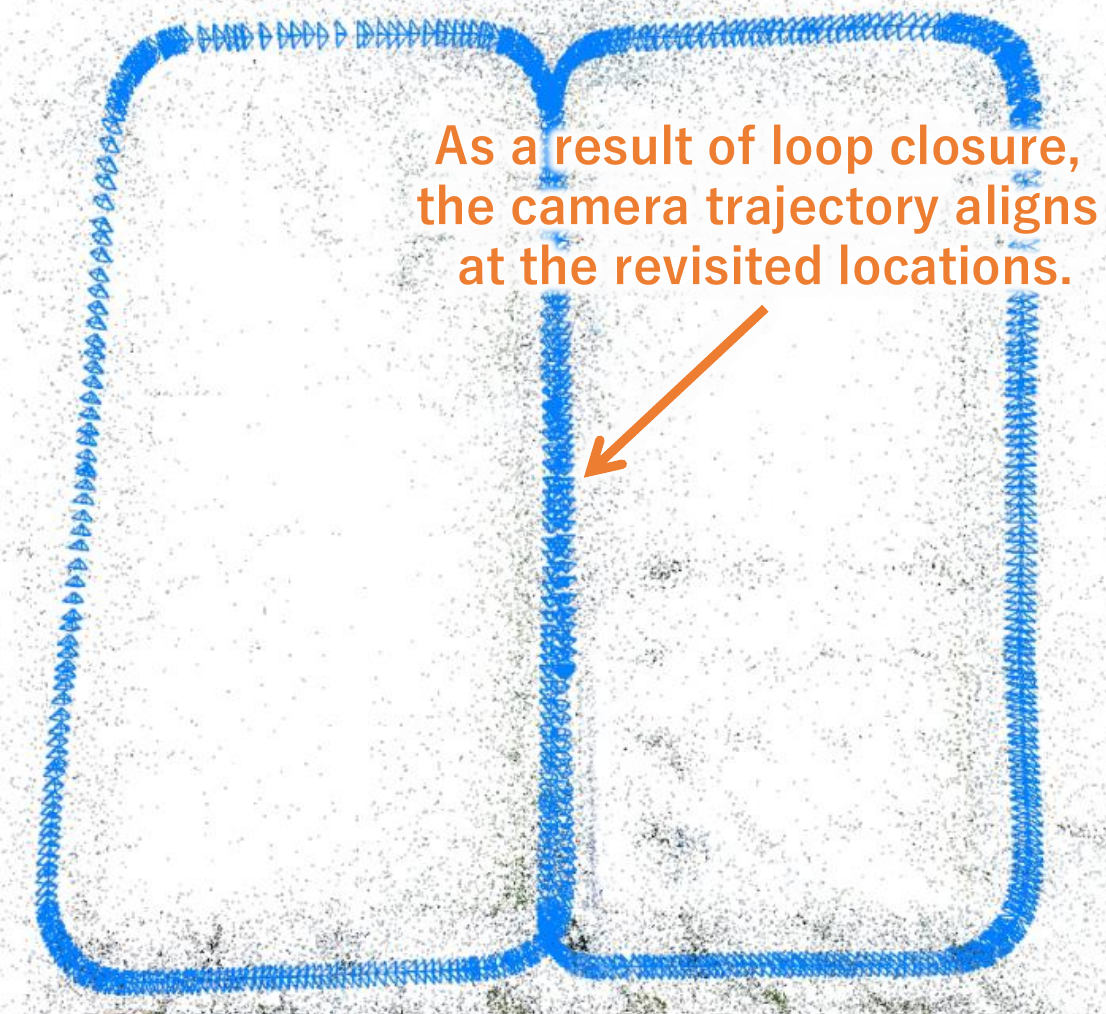}
    \caption{3D point cloud map and camera trajectory in the outdoor courtyard (proposed method).}
    \label{fig:experiment_outdoor2}
\end{figure}
\begin{figure}[t]
    \centering
    \includegraphics[width=0.9\linewidth]{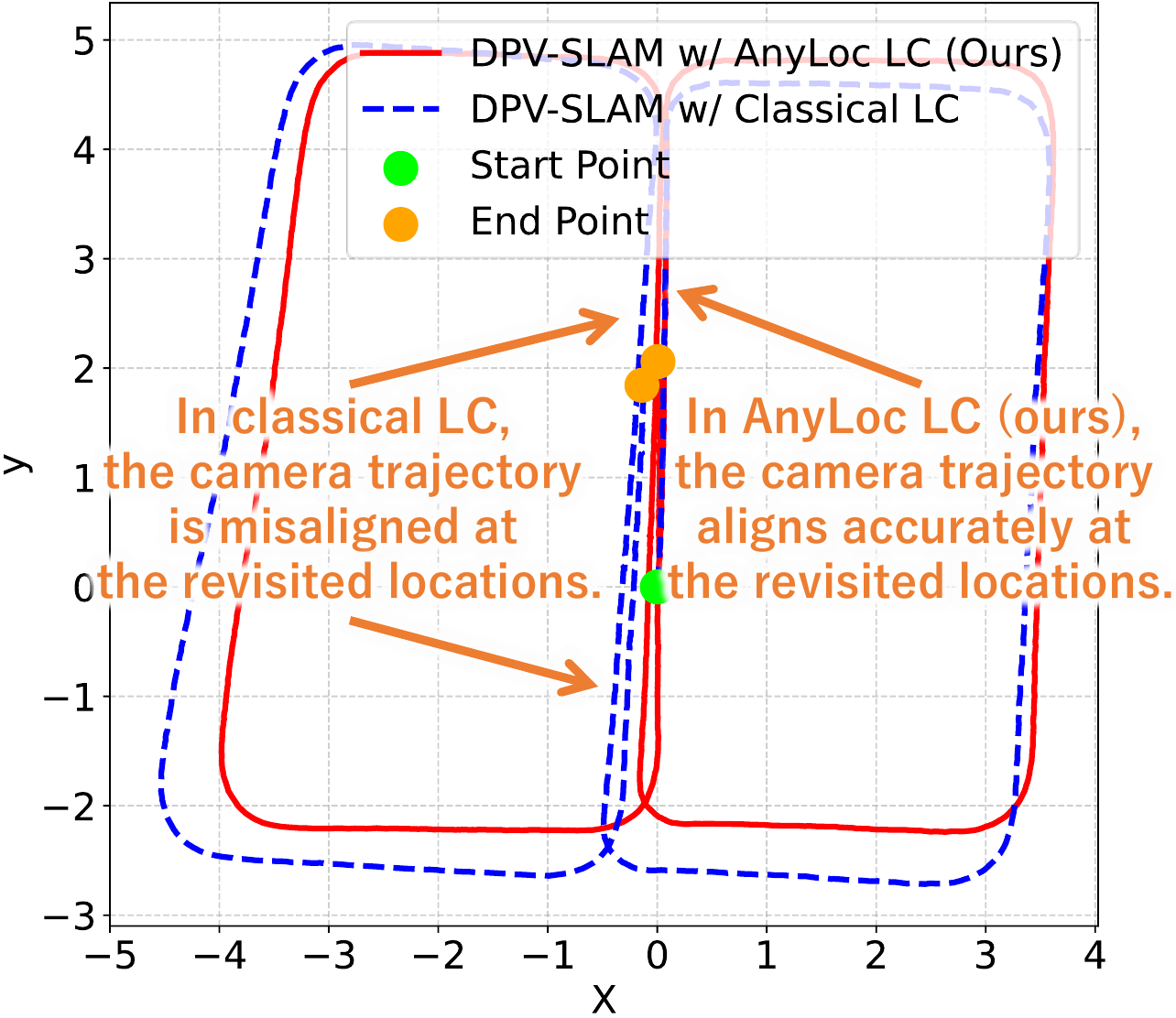}
    \caption{Comparison of camera trajectories in the outdoor environment (courtyard).}
    \label{fig:comparison_outdoor2}
\end{figure}

Fig.~\ref{fig:environment_outdoor2} shows the outdoor environment (courtyard) and its trajectory. The outdoor courtyard environment includes multiple sections along the path with similar textures such as grass and concrete. 

\begin{figure}[t]
    \centering
    \includegraphics[width=0.85\linewidth]{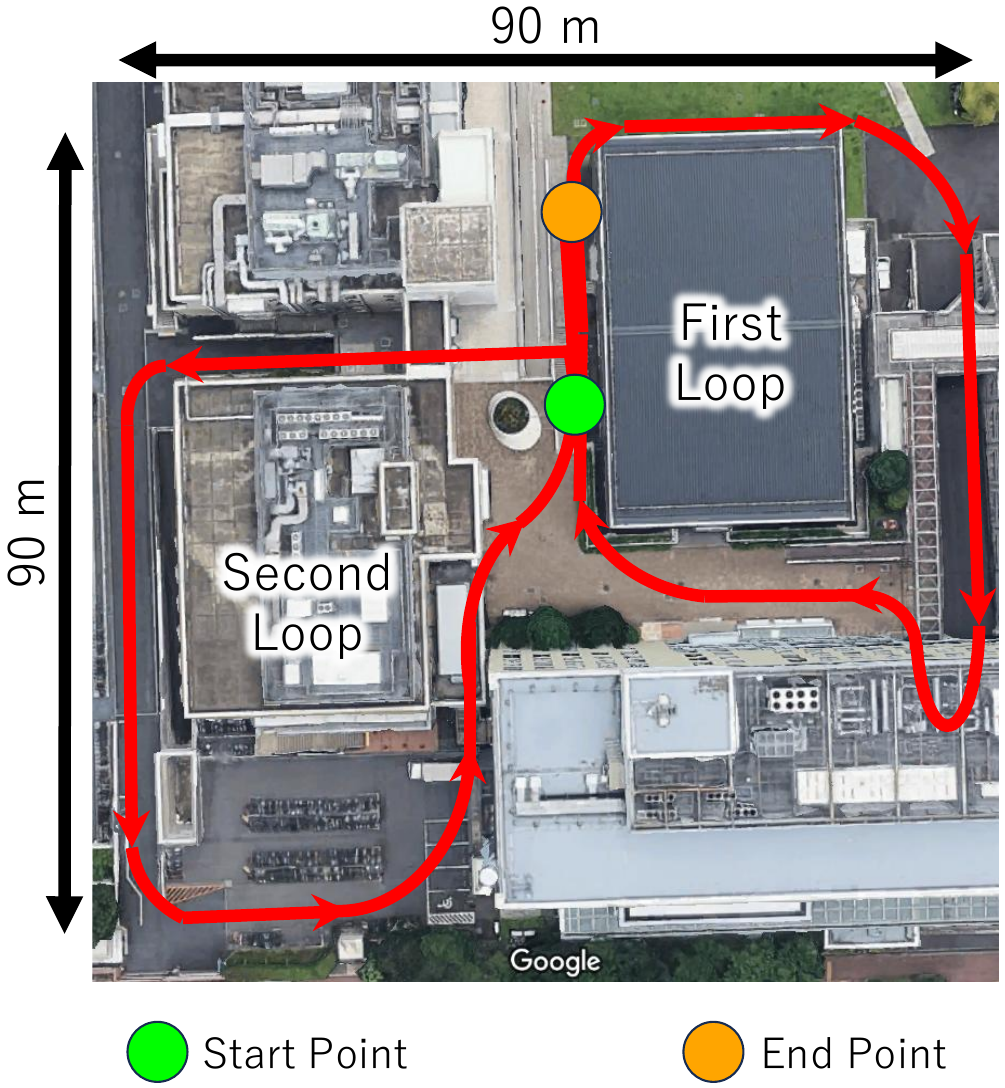}
    \caption{Outdoor environment (between buildings).}
    \label{fig:environment_outdoor1}
\end{figure}

Additionally, Fig.~\ref{fig:environment_outdoor1} shows the outdoor environment (between buildings) and its trajectory. This environment also contains several instances of pedestrian crossings.

% \subsection{Experiment Methodology}
% All data processing and algorithm execution were conducted on a desktop computer equipped with an NVIDIA GeForce GTX 4070 Ti SUPER GPU and an Intel i7-13700 CPU. 
\subsection{Data Collection and Processing}
A Logitech MX BRIO 700 (C1100) camera was used to capture images at a frame rate of 30 Hz. The camera was handheld during walking, and data was recorded using rosbag.

After data collection, experiments were conducted offline. The computer used for processing was a desktop running Ubuntu 22.04, equipped with an Intel Core i7-13700 CPU, 48 GB of RAM, and an NVIDIA GeForce RTX 4070 Ti SUPER GPU with 16 GB of VRAM.

\subsection{Evaluation Methodology}  
% 为了评估我们所提出模型的有效性，我们进行了定性的评估。评估的主要目标是将本方法与DPV-SLAM在回环检测精度、建图质量和整体系统性能方面进行比较分析。
% 在评估过程中，我们采用离线实验方式，使用统一的传感器采集数据，分别输入至所提出的方法与原始的 DPV-SLAM 系统中。通过对比两者在同一数据上的回环检测能力和建图质量，进一步验证本方法在实际场景下的表现优势。建图结果通过与真实环境的地图对齐误差进行评估，结合可视化对比，检验建图的完整性与一致性。
% 同时我们还对比了在三个环境中回环检测过程中的不同方法的图片检索情况，从而看出不同方法在闭环检测过程中的准确性的差异。
To evaluate the effectiveness of the proposed method, we conducted a qualitative assessment. The main objective of the evaluation is to compare our approach with DPV-SLAM in terms of loop detection accuracy, map reconstruction quality, and overall system performance.

During the evaluation, we adopted an offline experimental approach. Data were collected using a unified sensor setup and were separately processed by both the proposed method and the original DPV-SLAM system. By comparing the loop detection performance and map quality on the same dataset, we aimed to further validate the advantages of our method in practical scenarios. The mapping results were evaluated based on alignment errors with respect to the ground-truth environment map, and visual comparisons were also conducted to examine the completeness and consistency of the reconstructed maps.

Furthermore, we conducted a comparative analysis of image retrieval results produced by different loop detection methods in the three experimental environments, thereby revealing the differences in loop detection accuracy among the methods.

\subsection{Experimental Results}
% PS 3 Images
\begin{figure}[t]
    \centering
    \includegraphics[width=0.8\linewidth]{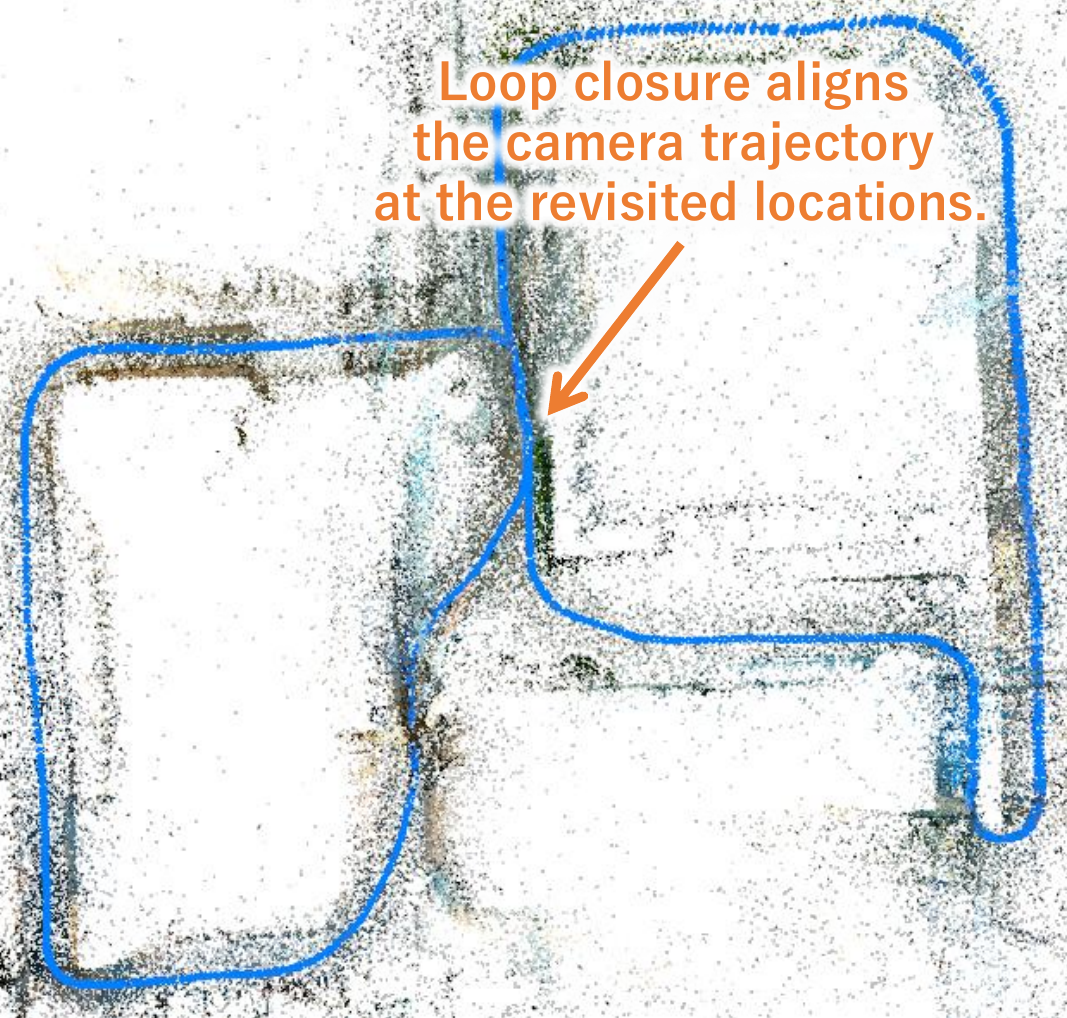}
    \caption{3D point cloud map and camera trajectory in the outdoor area between buildings (proposed method).}
    \label{fig:experiment_outdoor1}
\end{figure}
\begin{figure}[t]
    \centering
    \includegraphics[width=0.8\linewidth]{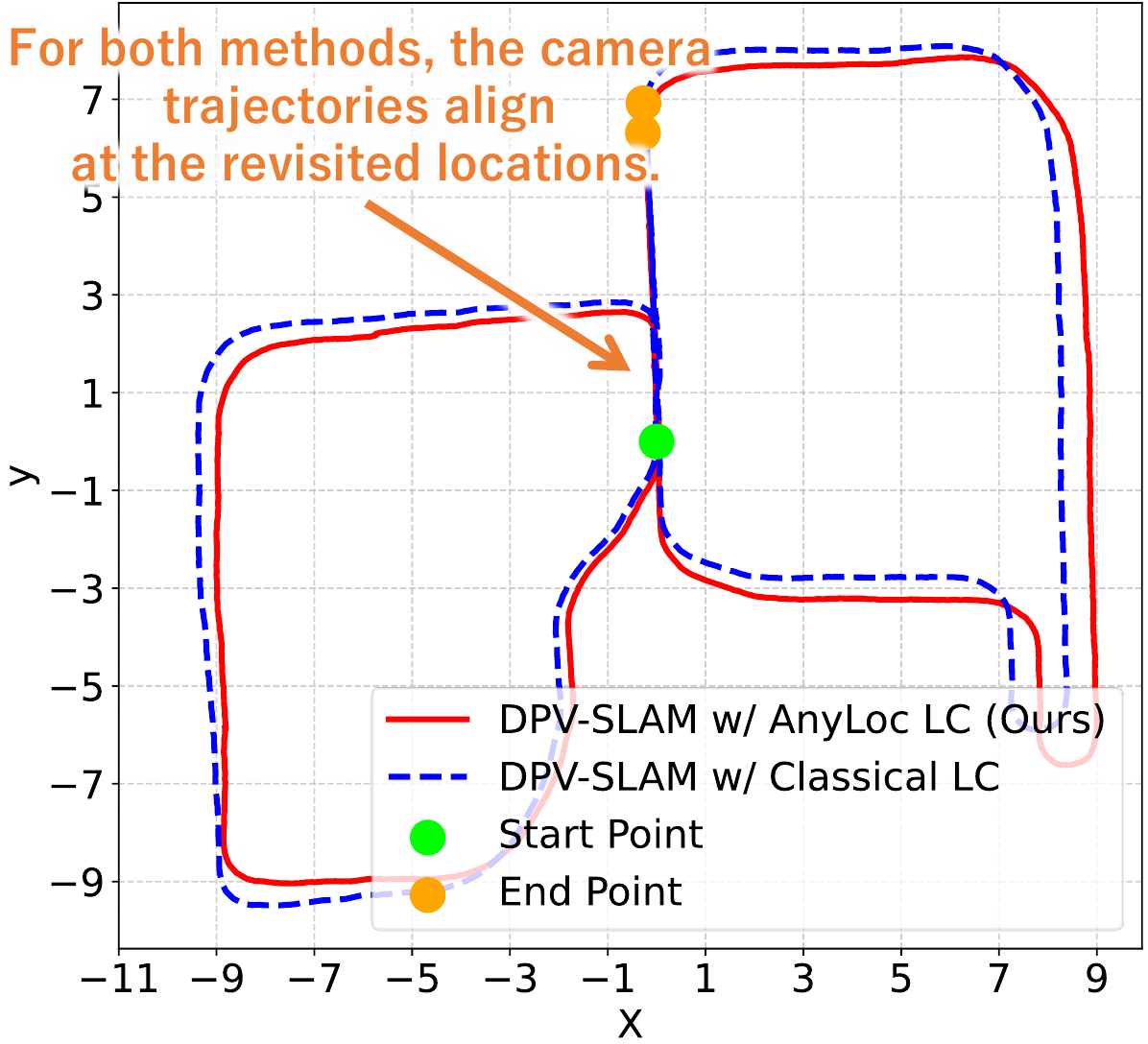}
    \caption{Comparison of camera trajectories in the outdoor environment (between buildings).}
    \label{fig:comparison_outdoor1}
\end{figure}

\begin{figure}[t]
% 上图显示了查询图像和匹配，两张图像检测了大量的配对点，我们标记出了其中的一部分。这些匹配点进行Keypoints Matching之后完成Pose graph optimization。
    \centering
    \includegraphics[width=0.95\linewidth]{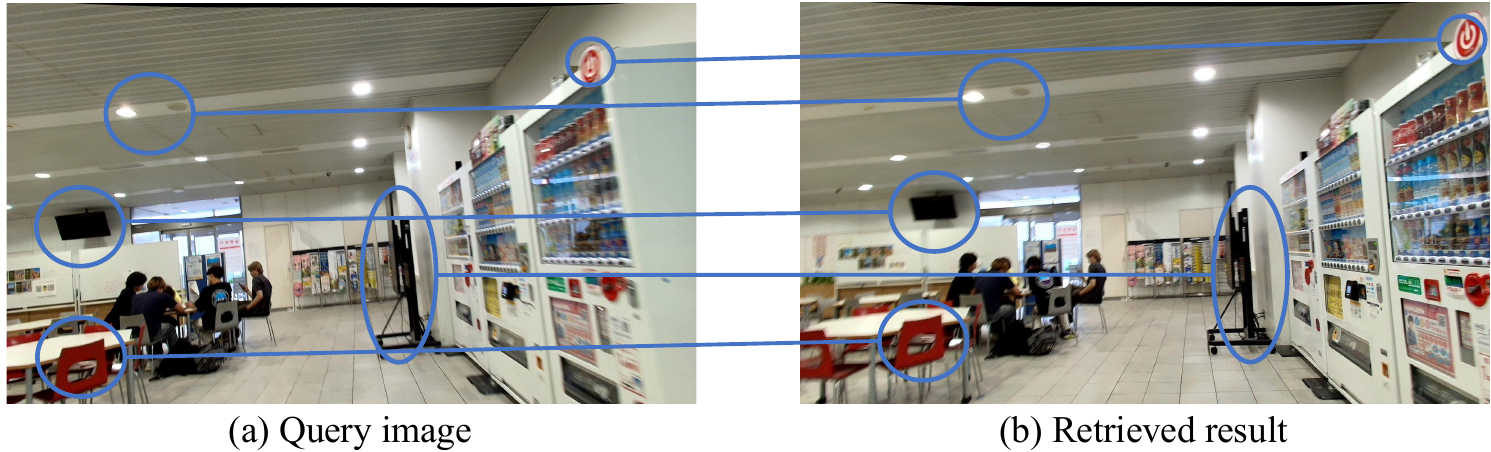}
    \caption{Loop closure retrieval performance (AnyLoc LC). A large number of correspondences are detected between the two images, and a subset of them is highlighted. These matched keypoints are subsequently used for registration and pose graph optimization.}
    \label{fig:DPV_AnyLoc_Retrieval}
\end{figure}
\begin{figure}[t]
% 显示了查询图像和匹配，我们将相似的特征区域标记了出来。虽然通过调整阈值获得了最相似的图像，但是获得的特征点不够多，没有通过几何验证。
    \centering
    \includegraphics[width=0.95\linewidth]{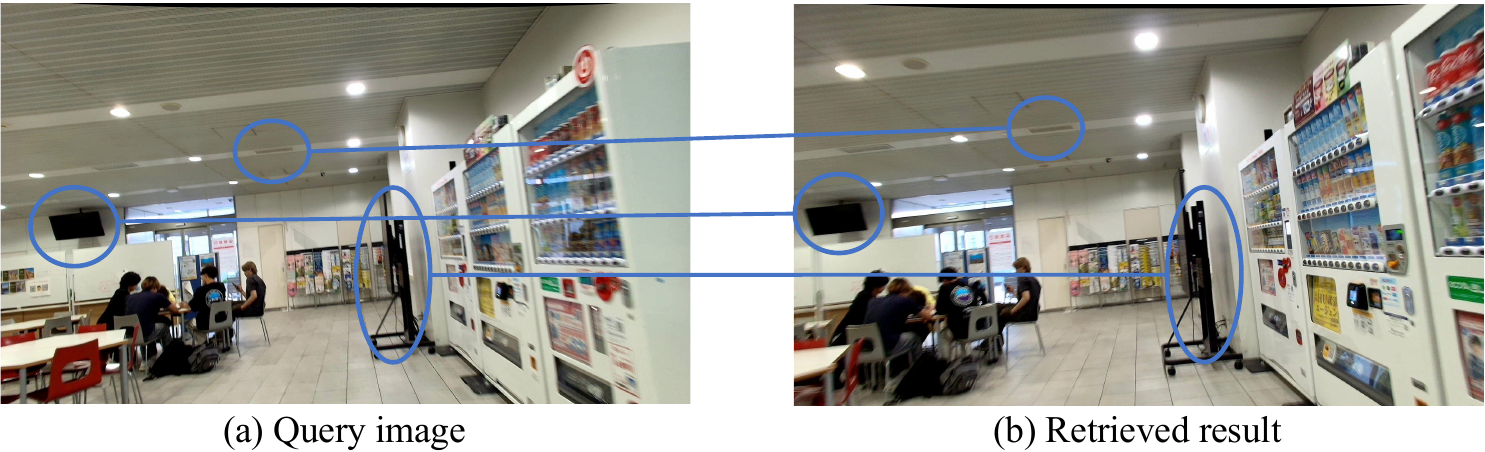}
    \caption{Loop closure retrieval performance (Classical LC). The figure displays the query image and its matches, with similar feature regions highlighted. Although the most similar images are obtained by adjusting the threshold, the number of matched keypoints is insufficient and fails geometric verification.}
    \label{fig:DPV_Retrieval}
\end{figure}
Fig.~\ref{fig:experiment_indoor} shows the 3D point cloud map and camera trajectory of the proposed method in the indoor environment. Fig.~\ref{fig:comparison_indoor} illustrates a comparison of camera trajectories in the same environment.

Next, Fig.~\ref{fig:experiment_outdoor2} shows the 3D point cloud map and camera trajectory of the proposed method in the outdoor environment (courtyard), while Fig.~\ref{fig:comparison_outdoor2} illustrates a comparison of camera trajectories.

%传统方法，即采用经典闭环检测（Classical Loop Closure, LC）的 DPV-SLAM，未能有效检测到有效的闭环，导致在回访路径上的相机轨迹出现显著偏差。相比之下，所提出的方法——结合 AnyLoc 闭环检测的 DPV-SLAM，成功检测到闭环，并保持了较高的位置精度，使得在闭环后回访位置的相机轨迹能够实现重叠。从结果来看，我们的方法有效地提高了闭环检测的准确率，从而使得得到的地图更加准确。这回答了\(\textbf{Q1}\).
The previous method failed to detect valid loops, resulting in significant misalignment of camera trajectories at revisited locations along the loop. In contrast, the proposed method successfully detected loops while maintaining high positional accuracy, causing the camera trajectories at revisited locations to overlap after loop closure. These results demonstrate that our method effectively improves the accuracy of loop detection, thereby leading to more precise map building. This addresses \hyperref[q1]{\textbf{Q1}}.

%接着，图~\ref{fig:experiment_outdoor1} 展示了室外环境（建筑物之间）的三维点云地图与相机轨迹，图~\ref{fig:comparison_outdoor1} 则对比了不同方法下的相机轨迹。在该环境中，传统的 DPV-SLAM（结合经典闭环检测 Classical LC）与所提出的结合 AnyLoc LC 的 DPV-SLAM 之间未观察到显著差异。两种方法均能正确检测到闭环，并在闭环后实现了回访位置处相机轨迹的重叠。
Then, Fig.~\ref{fig:experiment_outdoor1} shows the 3D point cloud map and camera trajectory in the outdoor environment (between buildings), while Fig.~\ref{fig:comparison_outdoor1} illustrates a comparison of camera trajectories. In this environment, no significant differences were observed between the conventional method and the proposed method. Both methods correctly detected loops, and loop closure resulted in overlapping camera trajectories at revisited locations.

% 另外，通过Fig.~\ref{fig:DPV_AnyLoc_Retrieval}可以看到我们提出的方法总能找到最相似的图片从而使其通过几何配对并完成闭环操作。相反，通过Fig.~\ref{fig:DPV_Retrieval}可以看到传统方法虽然检测到了闭环，但是检索的图片不能够通过最终的几何验证，从而不能实现闭环。这也反映了我们的方法在复杂环境下更为优秀。
Furthermore, as shown in Fig.\ref{fig:DPV_AnyLoc_Retrieval}, the proposed method consistently retrieves highly similar images that successfully pass geometric verification, enabling reliable loop closure. 
In contrast, as shown in Fig.\ref{fig:DPV_Retrieval}, although the classical method detects the loop closure, the retrieved image cannot pass the final geometric verification and thus the loop closure cannot be achieved. These results highlight the robustness and effectiveness of our method in challenging and complex environments.

\begin{figure}[t]
    \centering
    \footnotesize
    \tabcolsep=1mm
    \begin{tabular}{c}
    \includegraphics[width=0.9\linewidth]{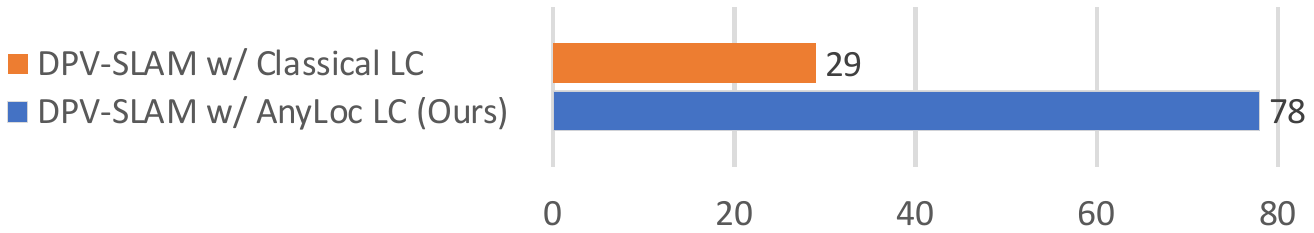} 
    \vspace*{1mm} \\
    (a) Corridor
    \vspace*{2mm} \\
    \includegraphics[width=0.9\linewidth]{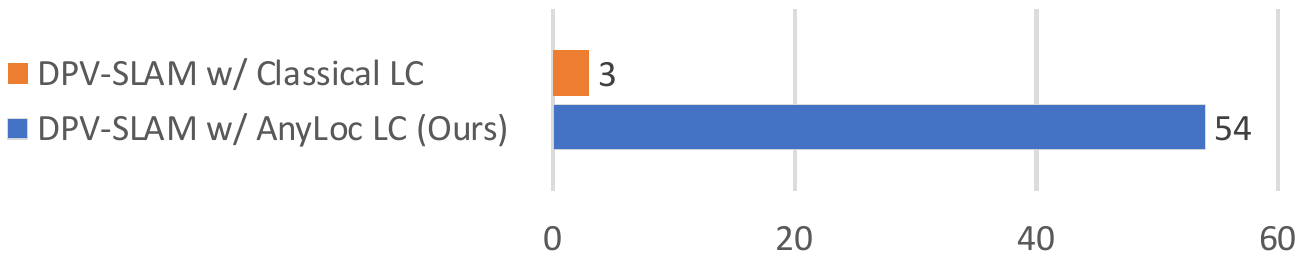} \vspace*{1mm} \\
    (b) Outdoor courtyard \vspace*{2mm} \\
    \includegraphics[width=0.9\linewidth]{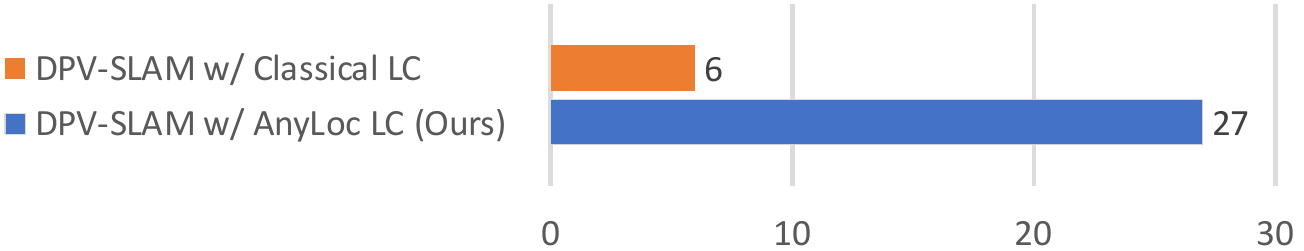} \vspace*{1mm} \\
    (c) Outdoor area between buildings \\
    \end{tabular}
    \caption{Comparison of the number of detected loops in different environments.}
    \label{fig:detected_matches_all}
\end{figure}

% 同时Fig.~\ref{fig:detected_matches_all}(a), Fig.~\ref{fig:detected_matches_all}(b), Fig.~\ref{fig:detected_matches_all}(c)显示了我们方法的闭环检测能力，这是我们的方法与设置为默认参数的传统方法的一次实验对比的结果,我们可以看到我们的方法在三种环境中相对于传统方法都能够检测到更多的闭环，通过调整threshold虽然可以增加检测数目，但是依然低于我们的方法。因此我们的方法大大地提升了整体的闭环检测的能力，进而提升生成地图的质量，这也是为什么上述的实验结果显示我们的方法生成的地图更加准确的原因。
% At the same time, as shown in Fig.\ref{fig:detected_matches_all}(a), Fig.\ref{fig:detected_matches_all}(b), and Fig.~\ref{fig:detected_matches_all}(c), we can see that our method can detect more loop closures than the traditional method in all three environments, thereby greatly improving the overall loop closure detection capability and further enhancing the quality of the generated map, which is also the reason why the above experimental results show that the maps generated by our method are more accurate.
As illustrated in Fig.~\ref{fig:detected_matches_all}(a)–(c), our method consistently identifies a greater number of loop closures than the conventional approach across all three environments. Although adjusting the detection threshold of the classical method can moderately increase its detection count, its performance remains inferior to ours. These observations indicate that the proposed method substantially enhances the robustness and overall capability of loop detection, thereby improving the quality and accuracy of the generated maps.

% 通过在室内、庭院和建筑物之间三种环境下的实验验证，所提出的方法在回环检测准确性和相机轨迹精度方面均优于传统的 DPV-SLAM。即使在光照不足、纹理重复或存在动态干扰的复杂场景中，本方法也表现出更高的鲁棒性与稳定性，具备良好的实际应用潜力，这回答了\(\textbf{Q2}\).
Through experiments in three different environments, the proposed method outperformed conventional DPV-SLAM in both loop detection accuracy and trajectory precision. Even under challenging conditions such as low lighting, repetitive textures, and dynamic interference, our method demonstrated higher robustness and stability, indicating strong potential for real-world deployment. This addresses \hyperref[q2]{\textbf{Q2}}.

\section{Conclusion}
% Ps ->Ours

% \begin{figure}[t]
%     \centering
%     \includegraphics[width=0.9\linewidth]{Comparison of the number of detected matches in corridor.pdf}
%     \caption{Comparison of the number of detected matches in corridor.}
%     \label{fig:detected matches in corridor}
% \end{figure}
% \begin{figure}[t]
%     \centering
%     \includegraphics[width=0.9\linewidth]{Comparison of the number of detected matches in courtyard.pdf}
%     \caption{Comparison of the number of detected matches in courtyard.}
%     \label{fig:detected matches in courtyard}
% \end{figure}
% \begin{figure}[!t]
%     \centering
%     \includegraphics[width=0.9\linewidth]{Comparison of the Number of Detected Matches on Roads Between Buildings.pdf}
%     \caption{Comparison of the number of detected matches on outdoor area\\ between buildings.}
%     \label{fig:detected matches on roads between buildings}
% \end{figure}
In this paper, we propose integrating AnyLoc, a visual place recognition method, into DPV-SLAM to replace the conventional Bag of Visual Words (BoVW) based loop detection. Furthermore, we develop a loop closure pipeline that combines adaptive similarity threshold adjustment with geometric verification, and evaluate its effectiveness through real-world experiments. Experimental results in real environments demonstrate a significant improvement in loop detection accuracy when combining AnyLoc with DPV-SLAM.

One limitation of the proposed method is its substantial requirement for GPU computational resources. Although the system can operate on CPUs, the processing speed decreases considerably.

In future work, we plan to improve loop detection speed by excluding unnecessary keyframe global descriptors using positional information. Additionally, we aim to evaluate our method on standard benchmark datasets.

\addtolength{\textheight}{-5cm}   % This command serves to balance the column lengths
                    
%%%%%%%%%%%%%%%%%%%%%%%%%%%%%%%%%%%%%%%%%%%%%%%%%%%%%%%%%%%%%%%%%%%%%%%%%%%%%%%%

\bibliographystyle{IEEEtran}
\bibliography{refer}

\end{document}